\documentclass{article}

\usepackage{arxiv}

\usepackage[utf8]{inputenc} 
\usepackage[T1]{fontenc}    
\usepackage{hyperref}       
\usepackage{url}            
\usepackage{booktabs}       
\usepackage{amsfonts}       
\usepackage{nicefrac}       
\usepackage{microtype}      
\usepackage{lipsum}
\usepackage{graphicx}
\usepackage{algorithm}
\usepackage{algorithmic}
\usepackage{array}
\usepackage{multirow}
\usepackage{threeparttable}

\usepackage{adjustbox}
\usepackage{amsmath}

\graphicspath{ {./images/} }

\title{Federated Hash Projected Latent Factor Learning}

\author{
 Jialan He \\
  College of Computer and Information Science\\
  Southwest University\\
  Chongqing 400715, China \\
  \texttt{nihhu2020@email.swu.edu.cn} \\
}

\renewcommand{\today}{}

\begin{document}
\maketitle
\begin{abstract}
Hash Learning (HL) is an efficient representation learning approach that maps real-valued data into compact binary representations. Traditional HL methods typically require users to upload personal data to a central server, which is incompatible with increasingly stringent data security regulations. Federated Learning (FL) provides a decentralized paradigm for learning globally optimal models without centralizing private data. However, most FL methods rely on transmitting large-scale real-valued gradient information, leading to high communication overhead and potential privacy risks. Integrating HL into FL is a promising solution. Nevertheless, existing HL methods suffer from limited representational capacity of binary codes, which may degrade model accuracy. To address this challenge, we propose a Federated Hash Projected Latent Factor (FHPLF) model. FHPLF introduces three key innovations: (a) replacing real-valued gradient matrices with binary gradient-like matrices, significantly reducing computation, storage, and communication costs while enhancing privacy protection; (b) leveraging Projected Hamming Distance for similarity modeling, which captures the importance of individual binary bits to improve representation capability; and (c) proposing a Secure Binary Gradient Reassembly and Privacy-Enhanced Upload (SBG-PEU) strategy to further reduce the risk of user interaction leakage during transmission. Extensive experiments on four real-world datasets demonstrate that FHPLF consistently outperforms state-of-the-art HL and FL methods, achieving a favorable trade-off among accuracy, efficiency, and privacy preservation.
\end{abstract}


\section{Introduction}
With the rapid expansion of digital information \cite{wu2026schemarag,ma2025review,tang2025auto}, users are increasingly confronted with information overload \cite{barathy2020applying,10977800,11077963}. Recommendation systems \cite{chronis2024survey} address this issue by filtering information and providing personalized content, services, and products. Among them, collaborative filtering \cite{wang2022hcfrec}, especially the Latent Factor (LF) model \cite{koren2009matrix, he2026survey, xu2025sampling, lyu2025genetic}, has been widely used for its accuracy and scalability \cite{wu20211, chen2024latent, yang2024latent, wang2025convolution}. LF-based methods \cite{lin2025neural,xu2025attention,lin2025momentum,mi2023spatio,zhong2023proximal} effectively address high-dimensional incomplete (HDI) data \cite{bi2023two,luo2020position,yuan2020multilayered} by learning compact latent representations of users and items in a shared latent space\cite{zhong2020momentum}, and predicting preferences via inner products. To further improve optimization efficiency, DPL \cite{lyu2025dynamic} introduces a dynamic PSO-based method for adaptive hyper-parameter tuning in latent factor analysis, enabling real-time parameter adjustment during training. Beyond recommendation, LFA-STSR \cite{wu2023robust} extends latent feature analysis with spatiotemporal regularization and L1-norm robustness for missing data recovery in WSNs under outliers. However, as data scale and dimensionality grow, repeated inner product computations introduce high computational and storage costs, limiting their efficiency in large-scale recommendation scenarios \cite{11201023,hansen2021projected}.

Hash Learning(HL) \cite{wang2017survey} offers an effective solution for improving recommendation efficiency. It encodes entity representations as binary codes in Hamming space, with each element set to +1 or -1, reducing storage from 64 bits to only one bit per element compared with real-valued representations \cite{yang2023label}. These compact binary codes also lower communication costs during representation transmission. Moreover, HL replaces costly inner product computations with efficient bitwise operations, leading to faster inference between entities \cite{zhang2016discrete,zhang2017discrete,liu2019compositional,hansen2020content}. Despite the advances of existing HL methods \cite{zhang2016discrete,zhang2017discrete,liu2019compositional,hansen2020content,hansen2021projected,liu2022hs}, they largely follow the centralized learning paradigm. Specifically, user-item interaction data are collected on a central server to construct the rating matrix for model training \cite{lin2020meta}, which requires users to disclose sensitive information and consequently raises privacy concerns \cite{gutierrez2024differential}.

Federated Learning (FL) \cite{guendouzi2023systematic} is a decentralized machine learning paradigm that enables collaborative model training without sharing raw user data \cite{ivannikova2019federated}. In FL, clients download the global model from a central server, update it using local data, and upload model parameters (e.g., gradients) for aggregation. The server then combines these updates to improve the global model through iterative communication rounds \cite{guendouzi2023systematic}. By keeping user data on local devices, FL effectively preserves privacy while maintaining model performance \cite{chai2020secure}. Owing to these advantages, FL has been widely adopted in representation learning, leading to various federated frameworks. A representative example is Federated Collaborative Filtering (FCF) \cite{ivannikova2019federated}, where user embeddings are updated locally, while item embeddings are jointly optimized by the server and clients.

Existing federated learning studies mainly operate in real-valued spaces, which face challenges in large-scale scenarios. Frequent communication between the server and numerous clients for model updates can impose substantial communication overhead \cite{rahimi2023evofed,zhang2023lightfr,al2023fedratrees,cui2022communication}. Moreover, the transmission of model gradients may expose sensitive client information, leading to potential privacy leakage \cite{zhu2019deep,wu2022federated,narula2024comprehensive}. To address these issues, \cite{zhang2023lightfr} introduced LightFR, which integrates HL with FL. Nevertheless, like most HL methods, LightFR relies on Hamming Distance for similarity measurement, which treats all bits equally and fails to model bit-level importance, leading to information loss and reduced representation accuracy. In contrast, Projected Hamming Distance mitigates this issue by projecting item binary representations into the query space to capture bit importance. Specifically, query bits with value -1 force corresponding item bits to -1, indicating irrelevance and emphasizing the remaining informative bits \cite{liu2022hs}.

To address the above inefficiencies and privacy concerns, this paper proposes a novel and efficient federated hashing learning framework, FHPLF. Designed to preserve client-side data privacy while enabling efficient and accurate binary representation learning, FHPLF provides several key advantages over existing methods \cite{ivannikova2019federated,chai2020secure,lin2020meta,lin2020fedrec,zhang2016discrete,zhang2017discrete,liu2019compositional,hansen2020content,hansen2021projected,liu2022hs,ijcai2023p507}. First, it enables globally optimal model learning without centralizing data, thereby ensuring strong privacy protection for local client information. Second, it incorporates Projected Hamming Distance \cite{liu2022hs} to assign different importance to bits, improving the expressiveness of binary representations. Third, its discrete binary representations significantly reduce communication, computation, and storage costs, while also mitigating the risk of gradient leakage \cite{chen2021wireless}. Specifically, FHPLF further includes an efficient federated discrete optimization strategy that transmits binary “gradient-like” signals to enhance communication efficiency and reduce privacy exposure, as well as a Secure Binary Gradient Reassembly and Privacy-Enhanced Upload (SBG-PEU) mechanism to prevent the leakage of user interaction information. Extensive experiments on four real-world datasets demonstrate that FHPLF achieves strong performance in privacy preservation, binary representation learning, and communication efficiency in federated learning settings.

\section{Related works}
\subsection{The Latent Factor Model}
The LF model \cite{xu2025adaptively, zhong2024alternating, wang2024distributed}, originally developed for recommender systems \cite{koren2009matrix}, has been widely adopted across various domains as an effective and scalable tool for representing high-dimensional incomplete (HDI) matrices \cite{li2025neural,yuan2025proportional,wu2024fine}, owing to its ability to capture latent structures underlying sparse and partially observed data\cite{li2023generalized,qin2023parallel}. Recent extensions primarily focus on improving data/structure modeling and training efficiency. For the data/structure modeling, PMLF \cite{9885025} densifies observations via prediction-sampling-based multilayer synthetic completion. And GLFA \cite{10179251} incorporates hidden graph structures and high-order interactions into recurrent factor learning. The nonnegativity-aware variants such as ADNLF and FNAE \cite{10159989,10265117} enhance expressiveness through adaptive divergence objectives or fast nonnegative autoencoder-style updates. More recently, MMA \cite{wu2026multimetric} proposes a multimetric autoencoder using multiple Lp-norms with adaptive weighting, enhancing representation learning for HDI data. MILFT-SSR \cite{yu2025multi} extends LF to multi-indicator tensor modeling for spatio-temporal signal recovery in WSNs, capturing inter-indicator and spatial correlations. EKL \cite{yuan2026novel} further incorporates Extended Kalman Filter-based temporal dynamics into latent feature analysis for time-varying QoS prediction. For the training efficiency, APNLF \cite{li2026adaptive} improves nonnegative latent factor learning by introducing a PID-incorporated optimization scheme with adaptive hyperparameter control, accelerating convergence of multiplicative updates. FPS \cite{10502217} introduces fuzzy PID-controlled SGD. PPL \cite{10380219} adopts pseudo-gradient-adjusted PSO for efficient hyperparameter adaptation, and MH \cite{9785520} leverages Hessian–vector products with momentum to enable efficient second-order optimization. Specialized designs further address temporal dynamics (KLFA) and QoS prediction robustness (D2E-LF) \cite{9839318,9783168}. Recently, federated extensions of LF have emerged to address privacy concerns in distributed scenarios. FLFT \cite{wu2026federated} extends latent factorization of tensors to a federated learning setting, enabling privacy-preserving representation learning over distributed tensor-structured interaction data. Similarly, FLFL \cite{yu2026federated} applies latent factor learning in a federated manner for spatio-temporal signal recovery in wireless sensor networks, where sensor observations are modeled as incomplete tensor-structured data and collaboratively recovered without sharing raw measurements. Overall, despite these advances, most LF variants still follow a centralized real-valued training paradigm that requires uploading or aggregating multi-party data at a server, incurring substantial computation/storage overhead at scale and introducing non-negligible privacy risks.

\subsection{Hash Learning}
HL \cite{wang2017survey} methods have made significant strides in reducing computational, storage, and communication costs by optimizing the binary representation of entities. \cite{zhang2016discrete} introduced Discrete Cooperative Filtering (DCF), a method that optimizes hash representations in a direct manner. Then, the same authors proposed Discrete Personalized Ranking (DPR) \cite{zhang2017discrete}, specifically tailored to handle implicit feedback. To amalgamate the strengths of discrete and real-valued representations, a hybrid collaborative filtering model known as Compositional Coding (CCCF) has been put forward \cite{liu2019compositional}. Deep hashing methods have been advanced by integrating them with various neural network architectures, such as Neu-hash \cite{hansen2020content} and HS-GCN \cite{liu2022hs}. \cite{hansen2021projected} introduced a novel model (VHPHD) based on Projection Hamming Distance to achieve more accurate binary representations. More recently, NGHF \cite{wu2026non} mitigates quantization loss by introducing a non-gradient discrete differential evolution framework for direct binary optimization, improving HDI representation accuracy. However, these methods rely on centralized algorithms and thus cannot safeguard the privacy of users' local data. \cite{zhang2023lightfr} proposed LightFR, which integrates HL with FL to protect local privacy and reduce communication costs. However, it overlooks the varying importance of bits within binary representations, resulting in information loss of binary representations.

\subsection{Federated Learning}
McMahan et al. \cite{mcmahan2017communication} introduced federated learning as a privacy-preserving machine learning paradigm, where users can contribute to model training without centralizing their raw data. Many researchers have integrated the concept of FL with efficient hidden feature models, developing numerous new approaches that safeguard user privacy while ensuring the predictive performance of machine learning models. FedCF \cite{ivannikova2019federated} pioneered the innovation of updating user embedding vectors locally at the client for implicit feedback data, with item embedding matrices being co-optimized between the server and client. Subsequently, FedMF \cite{chai2020secure} further enhanced privacy preservation. FedRec \cite{lin2020fedrec} is trained for explicit feedback data and combines FL with it. MetaMF \cite{lin2020meta} improves by reducing communication costs, and PFedRec \cite{ijcai2023p507} adds personalized learning to FedRec. FedDeepLF \cite{gao2025federated} further integrates deep neural networks with latent factor models to capture complex interactions and enhance privacy-preserving recommendation. All of the aforementioned models are trained in real-valued spaces and require the transmission of real gradient information, which can lead to high storage costs, longer computation times, and higher communication costs. Some studies have demonstrated that transmitting gradient information can expose models to gradient inversion and inference attacks \cite{zhu2019deep,tang2025fedckic}, leading to the leakage of local private data. In order to better protect data privacy, data encryption is required, and then FL will involve a large number of data encryption and decryption operations \cite{li2020review,banabilah2022federated,he2025fedai}.
However, it is computationally demanding; converting plaintext to ciphertext exponentially escalates computational and communication costs \cite{chen2021wireless,DBLP:series/synthesis/2019YangLCKCY}. Another solution is to adopt differential privacy, which achieves privacy protection by clipping the gradients or adding noise \cite{hu2025fgs,wu2025privacy}. However, this approach leads to significant gradient fluctuations and noise accumulation, which can affect the accuracy of the model.

\section{Preliminaries}

\textbf{Latent Factor Model.} 
Latent Factor (LF) models represent users and items in a shared low-dimensional latent space \cite{koren2009matrix}. Let
$x_u\in\mathbb{R}^{D}$ and
$y_i\in\mathbb{R}^{D}$
denote the latent vectors of user
$u\in U$
and item
$i\in I$,
respectively, where $D$ denotes the dimensionality of the latent feature space. The predicted rating$\hat{r}_{u,i}$ is estimated by the inner product of the corresponding latent vectors. The user and item latent matrices are defined as $X=[x_1,\ldots,x_m]\in\mathbb{R}^{|U|\times D}$,
$Y=[y_1,\ldots,y_n]\in\mathbb{R}^{|I|\times D}
$.

LF models learn these latent representations by minimizing the squared reconstruction error over the observed ratings. The optimization objective is formulated as
\begin{equation}
\label{eq1}
\mathop{\arg\min}_{X,Y}
\sum_{r_{u,i}\in\Lambda}
(r_{u,i}-\hat{r}_{u,i})^2
=
\mathop{\arg\min}_{X,Y}
\sum_{r_{u,i}\in\Lambda}
(r_{u,i}-x_u y_i^{\top})^2,
\end{equation}
where $\Lambda$ denotes the set of observed ratings in the user--item rating matrix
$\mathbf{R}\in\mathbb{R}^{m\times n}$.

\textbf{Federated Latent Factor Model.} We consider a federated recommendation scenario involving two types of entities: public entities (items) $i\in I$ and private entities (users) $u\in U$. Unlike conventional latent factor models, federated learning (FL) does not require the construction of a complete user--item rating matrix $\mathbf{R}\in\mathbb{R}^{m\times n}$ on a central server. Instead, each client corresponds to a user $u$ and locally stores the user's interaction data, including a set of interacted items $I_u$ and the corresponding ratings. The local interaction dataset of user $u$ is denoted as $\Omega_u={(i,r_{u,i}) \mid i\in I_u}$.

The learning objective remains consistent with that of the latent factor model, aiming to minimize the prediction error while preventing overfitting through regularization. The objective function is formulated as

\begin{equation}\label{eq2}
    \mathop{\arg\min}\limits_{X,Y} \sum_{r_{u,i}\in\Lambda} \left(r_{u,i} - x_u y_i^{\top}\right)^2+\lambda \left( \| x_u \|^2_2+ \| y_i \|^2_2 \right)
\end{equation}

where $\lambda$ denotes the regularization parameter.

Since the complete rating matrix is unavailable at the server, the training procedure must be adapted to the federated setting. Initially, each client initializes its local user latent vector $x_u$, while the server initializes the global item latent matrix $Y$. During each communication round, clients download the latest item latent matrix $Y$ from the server and compute the gradient of $x_u$ using their local interaction data $\Omega_u$. The user latent vector is updated as

\begin{equation}
\label{eq3}
\begin{aligned}
x_u^{t}
&\leftarrow
x_u^{t-1}
-\eta\nabla x_u,\
\nabla x_u
&=
\sum_{i\in I_u}
\left(
x_u^{\top}y_i-r_{u,i}
\right)y_i
+\lambda x_u,
\end{aligned}
\end{equation}

where $\eta$ denotes the learning rate.

After updating $x_u$, each client computes the gradients with respect to the item latent vectors and uploads them to the server. The server aggregates the received gradients to update the global item latent matrix. Specifically, the update rule for item latent vector $y_i$ is given by

\begin{equation}
\label{eq4}
\begin{aligned}
y_i^{t}
&\leftarrow
y_i^{t-1}
-\eta
\sum_{u\in U}
\nabla y_i^{u},\
\nabla y_i^{u}
&=
\left(
x_u^{\top}y_i-r_{u,i}
\right)x_u
+\lambda y_i.
\end{aligned}
\end{equation}

Through iterative client-side optimization and server-side aggregation, the latent representations of users and items are collaboratively learned without exposing users' raw interaction data.

\textbf{Projected Hamming Distance.}
Given a user $u$ and an item $i$, let $z_u,z_i\in\{-1,+1\}^{D}$ denote their binary hash codes in a $D$-dimensional Hamming space. Following \cite{hansen2021projected}, we use the Projected Hamming Distance (PHD) to measure the asymmetric discrepancy between $z_u$ and $z_i$. Specifically, the user code $z_u$ acts as a binary mask that projects the item code $z_i$ onto the dimensions considered important by the user.

\begin{equation}\label{eq5}
\begin{aligned}
   \delta_{PH}(z_u,z_i)
   &= \left\| z_u-P_{z_u}(z_i) \right\|_{H} \\
   &= \operatorname{SUM}\left(z_u \operatorname{XOR} 
   \underbrace{(z_u \operatorname{AND} z_i)}_{\text{projection}}\right) \\
   &= \operatorname{SUM}\left(z_u \operatorname{AND}(\operatorname{NOT} z_i)\right),
\end{aligned}
\end{equation}
where $z_u,z_i\in\{-1,+1\}^{D}$, and $P_{z_u}(z_i)=z_u\operatorname{AND}z_i$ denotes the projection of the item code $z_i$ onto the user code $z_u$. In this formulation, $z_u$ provides a query-specific binary importance mask: dimensions regarded as unimportant by the user are disabled, while only the active dimensions contribute to the dissimilarity. Therefore, PHD is asymmetric in general, i.e., $\delta_{PH}(z_u,z_i)\neq \delta_{PH}(z_i,z_u)$. In addition, the equivalent form $\operatorname{SUM}(z_u\operatorname{AND}(\operatorname{NOT}z_i))$ enables efficient computation, since $\operatorname{NOT}z_i$ can be pre-computed and cached for static item codes.

\section{The proposed FHPLF framework}

\subsection{Problem Definition}
Like the conventional LF model, FHPLF aims to reconstruct the observed ratings in the user--item rating matrix $R$. The key difference is that FHPLF performs rating prediction in the binary hamming space rather than the real-valued latent space. Based on the Projected Hamming Distance defined in Eq. \eqref{eq5}, the predicted rating of user $u$ on item $i$, denoted by $\hat{r}_{u,i}$, is formulated as:

\begin{equation}\label{eq6}
\begin{aligned}
  \hat{r}_{u,i}&= \frac{1}{D}\left(D-\delta_{PH}\left(w_u,q_i\right)\right) \\
 & =1-\frac{1}{D}  ( \frac{w_u+1}{2} )^\top  ( \frac{-q_i+1}{2} )  \\
  &= \frac{3}{4} +\frac{1}{4D} ( w_u^\top q_i - \sum_{d}^{D}w_{u,d} + \sum_{d}^{D}q_{i,d})   \\
\end{aligned}
\end{equation}
\begin{equation*}
s.t. w_u \in \{+1, -1\}^D, q_i \in \{+1, -1\}^D
\end{equation*}
where $w_{u,d}$ and $q_{i,d}$ denote the $d$-th bit of the binary representations $w_u$ and $q_i$, respectively. Similar to the LF model, FHPLF aims to minimize the prediction error between $\hat{r}_{u,i}$ and $r_{u,i}$. Unlike LF, FHPLF operates in the binary hamming space and estimates user preferences using PHD. By leveraging efficient bitwise operations, FHPLF reduces memory consumption and accelerates inference.

Following prior studies \cite{zhang2016discrete}, balanced binary codes encourage each bit to take +1 and -1 with equal probability, thereby maximizing the entropy of individual bits. Therefore, in addition to the binary constraints, we impose balance constraints on the user and item representations. The objective function of FHPLF is formulated as:
\begin{align}\label{eq7}
\nonumber
 &\mathop{\arg\min}\limits_{W,Q} \sum_{u \in U }  \sum_{i \in I_u} ( r_{u,i} \! - \! (\frac{3}{4} \!+\!\frac{1}{4D}( w_u^\top q_i \! -\! \sum_{d}w_u \!+\! \sum_{d}q_i)))^2  \\ 
  \nonumber
 &s.t. \underbrace{(w_u \in \{+1, -1\}^D, q_i \in \{+1, -1\}^D)}_{\text{Binary Constraints}} \\ 
 & s.t. \underbrace{\sum_{d=1}^{D}w_{u,d}=0, \sum_{d=1}^{D}q_{i,d}=0}_{\text{Balanced  Constraints}}
\end{align}
where $w_u$ and $q_i$ denote the $u$-th and $i$-th row vectors of the binary representation matrices  $W\in\{+1,-1\}^{|U|\times D}$ and $Q\in\{+1,-1\}^{|I|\times D}$, respectively.

However, Eq. \eqref{eq7} defines a discrete optimization problem with binary and balance constraints, which is NP-hard \cite{zhou2012learning,zhang2016discrete}. To make the optimization tractable, we relax the balance constraints using a penalty-based formulation. Specifically, Eq. \eqref{eq7} is reformulated as:

\begin{equation}\label{eq8}
\begin{aligned}
 \underset{\substack{W, Q}}{\arg \min } \sum_{\substack{u \in U \\ i \in I_{u}}}& (r_{u, i}-(w_{u}^{T} q_{i}-\sum_{d} w_{u}+\sum_{d} q_{i}))^{2} \\
&+\alpha\|\sum_{d} w_{u}\|^{2}+\beta\|\sum_{d} q_{i}\|^{2}
\end{aligned}
\end{equation}
where $r_{u,i}$ denotes the original true rating with $r_{u,i}\in \left[0, 1 \right]$, and we scaled the $r_{u,i}\! \gets \! 4Dr_{u,i} \! -\! 3D$. The first term in this formula represents the loss of the observed ratings $r_{u,i}$, while the second term indicates our preference for balanced binary representations. Here, $\alpha$ and $\beta$ are the regularization controlling hyperparameters that control the trade-off between minimizing the empirical error and implementing the balanced constraints. $\left \| \cdot \right \| $ indicates the Euclidean norm of a vector, following the previous studies \cite{zhou2012learning,zhang2016discrete}. Such a regularization term in Eq. \eqref{eq8} not only maximizes the information entropy of the bit but also helps to avoid overfitting. Next, we will elaborate on the optimization of FHPLF in detail. 

\subsection{Model Optimization}

\begin{figure}
	\centering
	\includegraphics[width=0.9\columnwidth]{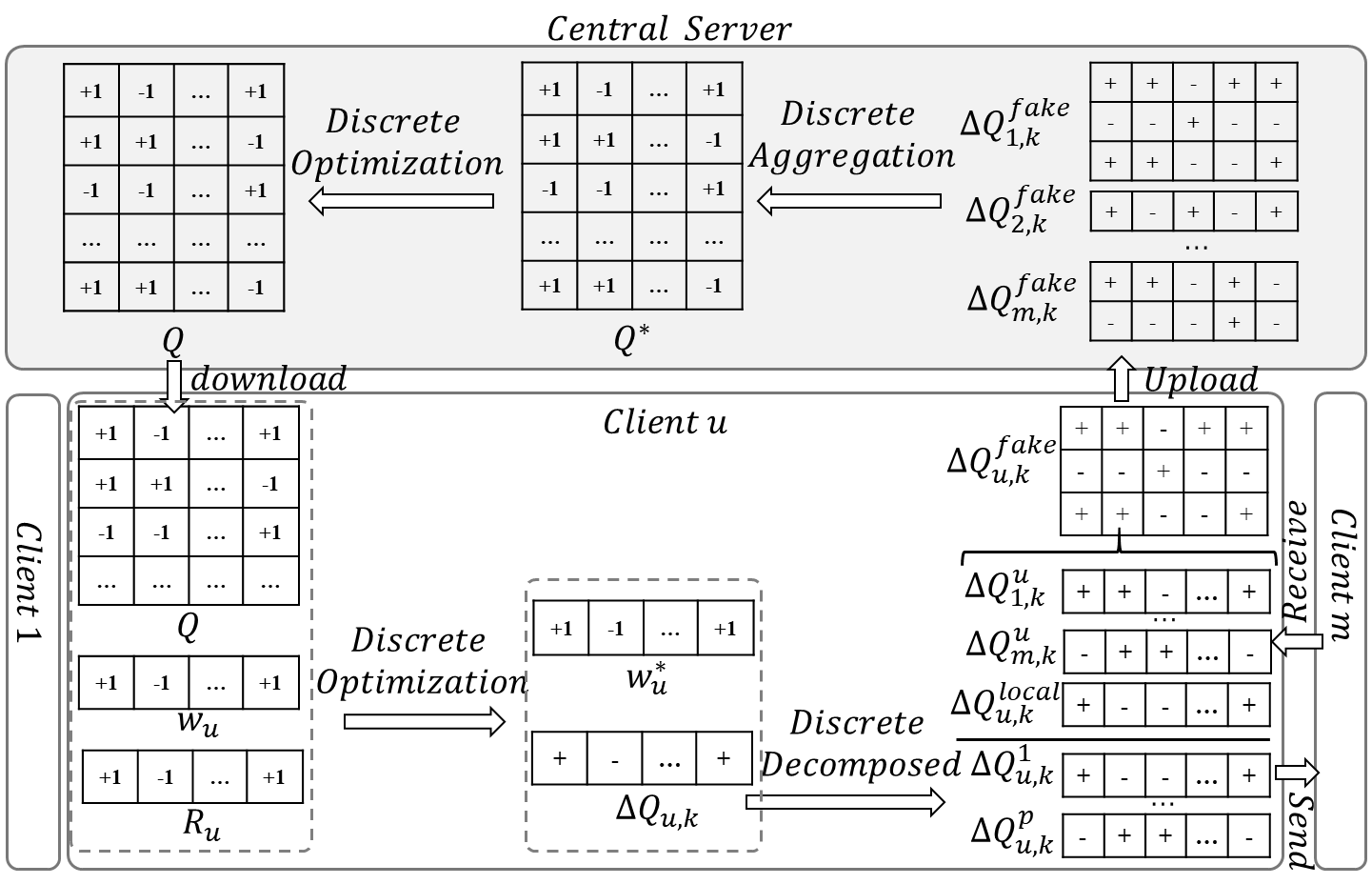}
	\caption{Overview of the proposed FHPLF framework.}
	\label{fig:fig1}
\end{figure}

Fig. \ref{fig:fig1} illustrates the training procedure of FHPLF. At the beginning, the central server randomly initializes the item binary representation matrix $Q \in \{+1,-1\}^{|I|\times D}$, while each client initializes its local user binary vector $w_u \in \{+1,-1\}^{D}$. At each communication round $r$, the server broadcasts the current item representation matrix $Q$ to all clients. Each client then updates its private user representation $w_u$ based on its local dataset $\Omega_u$, and computes a "gradient-like" update $\Delta Q_{u,k}$ for the item representations via discrete optimization. To enhance privacy preservation, $\Delta Q_{u,k}$ is decomposed into $P_u$ fragments, denoted as $\{\Delta Q^p_{u,k}\}_{p=1}^{P_u}$. Each client retains one fragment locally and exchanges the remaining fragments with $P_u-1$ other clients. The uploaded update $\Delta Q^{\text{fake}}_{u,k}$ is constructed by aggregating the received fragments together with the local fragment. The central server collects $\Delta Q^{\text{fake}}_{u,k}$ from all clients and performs global aggregation to update $Q$, which is then broadcast for the next round of optimization. This federated discrete optimization process is repeated until convergence.

Firstly, the central server initializes the binary representations of public entities, denoted as $Q \in \{+1,-1\}^{|I|\times D}$, while each client initializes the local binary representation $w_u \in \{+1,-1\}^{D}$ for user $u$. The training procedure can then be divided into two main stages: \textbf{local update} and \textbf{global update}.

\subsubsection{Local Update}
At the beginning of each communication round, each client downloads the current public entity representation matrix $Q$ from the server to the local environment. Then, the user-specific binary representation $w_u$ is updated locally using the private dataset $\Omega_u$. This process is essential for preserving the confidentiality of local data.

Since the objective function in Eq. \eqref{eq8} is decomposable over users, the binary representation $w_u$ for user $u$ can be independently updated at each client according to the following formulation:

\begin{equation}\label{eq9}
\begin{aligned}
&\underset{w_u}{\arg\min} \sum_{i \in I_u} (r_{u,i} - \hat{r}_{u,i})^2 + \alpha \| \sum_d w_u \|^2 \\
& \hat{r}_{u,i} = w_u^T q_i - \sum_d w_u + \sum_d q_i
\end{aligned}
\end{equation}

We adopt a discrete optimization strategy, namely the Discrete Coordinate Descent (DCD) algorithm \cite{zhang2016discrete}, to learn $w_u$ bit by bit. Let $w_{u,k}$ denote the $k$-th bit of the binary representation $w_u$, and let $w_{u,k’}$ denote the remaining bits excluding the $k$-th component. Accordingly, we decompose $w_u$ and $q_i$ as $w_u = [w_{u,k}, w_{u,k’}]$ and $q_i = [q_{i,k}, q_{i,k’}]$, respectively. Based on this decomposition, we can obtain:
\begin{equation}\label{eq10}
\begin{aligned}
\begin{cases}
\sum_d w_u = \sum w_{u,k'} + w_{u,k} \\
\sum_d q_i = \sum q_{i,k'} + q_{i,k} \\
w_u^T q_i = w_{u,k'}^T q_{i,k'} + w_{u,k} q_{i,k}
\end{cases}
\end{aligned}
\end{equation}

We fix $w_{u,k’}$ and $q_i$, and update the $k$-th bit $w_{u,k}$ according to Eq. \eqref{eq9} and Eq. \eqref{eq10}. Accordingly, the update of $w_u$ at each local client can be formulated as the following optimization problem:
\begin{equation}\label{eq11}
\begin{aligned}
\underset{w_u}{\arg\min} -w_{u,k} w_{u,k}^{*}
\end{aligned}
\end{equation}
where $w^{*}_{u,k}$ serves as the guiding factor for determining the update direction of $w_{u,k}$. We then solve the $w^{*}_{u,k}$ by using the following equations: 
\begin{equation}\label{eq12}
\begin{aligned}
\Delta w_{u,k}^{i}
=(r_{u,i}-\tilde r_{u,i})(q_{i,k}-1)
\end{aligned}
\end{equation}
\begin{equation}\label{eq13}
\Delta w_{u,k} = \sum_{i \in I_u} \Delta w_{u,k}^i
\end{equation}
\begin{equation}\label{eq14}
w_{u,k}^* = \Delta w_{u,k} - \alpha \sum w_{u,k'}
\end{equation}
where $\tilde r_{u,i}= w_{u,k'}^{\mathsf T} q_{i,k'} - \sum w_{u,k'} + \sum q_{i,k'} - 1$, and $\Delta w_{u,k}$ denotes a "gradient-like" term that plays a role similar to the gradient in traditional gradient descent optimization. Specifically, $\Delta w^{i}_{u,k}$ represents the contribution of item $i$ to the gradient-like update of the $k$-th bit of user $u$, where $i \in I_u$.

Following Eq. \eqref{eq11}, we fix $w_{u,k’}$ and $q_i$ to update the $k$-th bit $w_{u,k}$. When $w^*_{u,k} = 0$, the value of $w_{u,k}$ remains unchanged during the update. Otherwise, if $w^*_{u,k} \neq 0$, the optimal solution for $w_{u,k}$ is given by the sign of $w^*_{u,k}$. Therefore, we update $w_u$ in a bit-wise manner according to the following protocol:
\begin{equation}\label{eq15}
w_{u,k} =
\begin{cases}
\mathrm{sign}(w_{u,k}), & \text{if } w_{u,k}^* = 0 \\
\mathrm{sign}(w_{u,k}^*), & \text{if } w_{u,k}^* \neq 0
\end{cases}
\end{equation}
where $\mathrm{sign}(\cdot)$ denotes the sign function. Thus, the binary representation $w_u$ is updated iteratively using locally available data until convergence, which is achieved when all bits become stable.

\subsubsection{Global Update}
The central server broadcasts a signal along with the current binary representation of the public entity to the corresponding clients before updating $q_i$.

Subsequently, each client receives the control signal along with the current binary representation of the public entity, and uploads the necessary model information for updating $q_i$. Based on Eq. \eqref{eq8}, the central server updates $q_i$ according to the following equation:
\begin{equation}\label{eq16}
\begin{aligned}
&\underset{q_i}{\arg \min} \sum_{u \in U} \left(r_{u,i} - \hat{r}_{u,i}\right)^2 + \beta \| \sum_d q_i \|^2 \\
&\hat{r}_{u,i} = w_u^T q_i - \sum_d w_u + \sum_d q_i
\end{aligned}
\end{equation}

For the global update, we adopt the Discrete Coordinate Descent (DCD) algorithm to iteratively update each bit of $q_i$. Let $q_{i,k}$ denote the $k$-th bit of the public entity representation $q_i$, and let $q_{i,k’}$ denote the remaining bits excluding the $k$-th component.

Thus, we assume $q_i = [q_{i,k}, q_{i,k’}]$ and $w_u = [w_{u,k}, w_{u,k’}]$. We fix $q_{i,k’}$ and $w_u$ to update $q_{i,k}$ according to Eq. \eqref{eq9} and Eq. \eqref{eq10}. Accordingly, $q_{i,k}$ can be updated by solving the following optimization problem:
\begin{equation}\label{eq17}
\underset{W}{\arg \min} \ - q_{i,k} q_{i,k}^{*}
\end{equation}
where $q^*_{i,k}$ serves as the guiding factor for determining the update direction of $q_{i,k}$. In order to get the $q^*_{i,k}$, each client need to solve the following problem:
\begin{equation}\label{eq18}
\begin{aligned}
&\Delta q_{i,k}^{u}
=(r_{u,i}
-\tilde r_{u,i})(w_{u,k}+1)
\end{aligned}
\end{equation}
where $\Delta q^u_{i,k}$ represents a "gradient-like" term associated with the $k$-th bit of the binary vector $q_i$, which is locally computed by client $u$. To enhance storage and communication efficiency, as well as ensure privacy preservation, each local client uses $\mathrm{sign}(\cdot)$ to generate $\Delta \hat{q}^u_{i,k}$ in the binary space as in Eq. \eqref{eq14}, and then uploads $\Delta \hat{q}^u_{i,k}$ to the central server.

\begin{equation}\label{eq19}
\begin{aligned}
\Delta \hat{q}_{i,k}^{u} = \mathrm{sign}( \Delta q_{i,k}^{u})
\end{aligned}
\end{equation}
Subsequently, the central server performs aggregation to obtain $q^*_{i,k}$:
\begin{equation}\label{eq20}
\begin{aligned}
q_{i,k}^{*} = \sum_{u \in U} \Delta \hat{q}_{i,k}^{u} - \beta \sum q_{i,k'}
\end{aligned}
\end{equation}
where $q^*_{i,k}$ serves as the guiding factor for determining the update direction of $q_i$. According to the objective function in Eq. \eqref{eq16}, the server can update $q_{i,k}$ by the following equation:
\vspace{-10pt}
\begin{equation}\label{eq21}
q_{i,k} = \begin{cases} 
\mathrm{sign}(q_{i,k}),& \text{if } q_{i,k}^{*} = 0 \\
\mathrm{sign}(q_{i,k}^{*}),& \text{if } q_{i,k}^{*} \neq 0
\end{cases}
\end{equation}

To accelerate the training process, on each client, we perform the computation of $\Delta \hat{q}^u_{1,k}$, $\Delta \hat{q}^u_{2,k}$, …, $\Delta \hat{q}^u_{|I_u|,k}$ in parallel. Specifically, we define $\Delta Q_{u,k} = [\Delta \hat{q}^u_{1,k}, \Delta \hat{q}^u_{2,k}, …, \Delta \hat{q}^u_{|I_u|,k}]$ as  the "gradient-like" vector, which originating from local client $u$, based on the local data $(u, i, r_{u,i}|i \in I_u) \in \Omega_u$. 

Additionally, we define $\Delta Q_k$ as the final global "gradient-like" vector, which is constructed from the $k$-th representation "gradient-like" terms $\Delta \hat{q}^u_{i,k}$ $k$-th of all items representations as following equation: 
\begin{equation}\label{eq22}
\begin{aligned}
\Delta Q_k = \mathrm{Agg}\left( \left\{ \Delta Q_{1,k}, \dots, \Delta Q_{u,k}, \dots, \Delta Q_{|U|,k} \right\} \right)
\end{aligned}
\end{equation}
where $ \mathrm{Agg}(\cdot)$ is an discrete aggregation function that sums the "gradient-like" terms $\Delta \hat{q}^u_{i,k}$ for each item $i$ across all clients, where these terms are contributed by uploading from each client.
\begin{equation}\label{eq23}
\begin{aligned}
Q_k^{*} = \Delta Q_k - \beta \sum Q_{k'} 
\end{aligned}
\end{equation}
where $Q^*_k= [q^*_{1,k}, q^*_{2,k}, …, q^*_{|I|,k}]$ is the final guiding factors matrix. Considering Eq. \eqref{eq12}, which indicates that the signs of $q_{i,k}$ and $q^*_{i,k}$ should be the same to minimize the objective function, we use the following formula to update the binary representation matrix of the public entities:
\begin{equation}\label{eq24}
\begin{aligned}
Q_k = \mathrm{sign}\left( F\left( Q_k, Q_k^{*} \right) \right)
\end{aligned}
\end{equation}

According to Eq. \eqref{eq23} once the central server has updated $Q_k$ the client re-downloads $Q_k$ and then proceeds with the optimization of the subsequent bit, $Q_{k+1}$. 

Consequently, the binary representation matrix $Q$ can be updated iteratively in a bit-wise manner. This process does not require uploading locally client data to the central server and continues until convergence, which is achieved when all bits in $Q$ remain stable.

It is important to note that the "gradient-like" matrix $\Delta Q_{u,k}$ uploaded by each client is binary, which significantly reduces communication overhead. Compared with traditional real-valued gradients, this binary transmission provides several advantages in federated learning. In particular, it mitigates the gradient leakage issue associated with real-valued updates.

\subsection{SBG-PEU Strategy}
For each client, only items $i \in I_u$ have non-zero "gradient-like" terms for updating their binary representations. Therefore, an attacker can easily infer the complete user--item interaction history from these non-zero "gradient-like" signals \cite{wu2022federated}. To address this issue, we propose a novel strategy, namely Secure Binary Gradient Reassembly and Privacy-Enhanced Upload (SBG-PEU).

After the client $u$ completes the computation of the "gradient-like" vector $\Delta Q_{u,k} = [\Delta \hat{q}^u_{1,k}, \Delta \hat{q}^u_{2,k}, …, \Delta \hat{q}^u_{|I_u|,k}]$, it is essential to perform a discrete decomposition of this vector. Thus, generate $P$ binary vectors ${\Delta Q^1_{u,k},…, \Delta Q^P_{u,k}}$ randomly, each termed a fragment of the "gradient-like" vector $\Delta Q_{u,k}$. These vectors must satisfy certain constraints, as shown in Eq. \eqref{eq25}:
\vspace{-10pt}
\begin{equation}\label{eq25}
\begin{aligned}
\sum_{p=1}^{P} \Delta Q_{u,k}^{p} = \Delta Q_{u,k}
\end{aligned}
\end{equation}
Then, $P-1$ clients are randomly selected to receive the fragments, while each client retains one fragment locally. After all clients complete the exchange of "gradient-like" vector fragments, each client aggregates the received fragments together with its local fragment to construct a fake gradient matrix, which is then uploaded to the central server, as shown in Eq. \eqref{eq26}.
\begin{equation}\label{eq26}
\begin{aligned}
\Delta Q_{u,k}^{fake} = \left\{  \Delta Q_{u,k}^{local},\ \smash[b]{\Delta Q_{1,k}^{u}, \dots, \Delta Q_{r,k}^{u} , \dots,  \Delta Q_{|U|,k}^{u}} \right\}
\end{aligned}
\end{equation}
where $\Delta Q^u_{r,k}$ indicates a fragment that client $u$ has received from other clients. 

Since the central server updates based solely on the aggregated value post-decomposition, the update direction remains consistent with that of the original "gradient-like" matrix upload. It is easy to prove that Eq. \eqref{eq21} is equal to the following equation:
\begin{equation}\label{eq27}
\begin{aligned}
\Delta Q_k^{fake} \!=\!  \mathrm{Agg} \! \left(\! \left\{ \Delta Q_{1,\text{k}}^{fake}, \dots, \Delta Q_{u,\text{k}}^{fake}, \dots, \Delta Q_{|U|,\text{k}}^{fake} \right\}\! \right)
\end{aligned}
\end{equation}

By decomposing and fragmenting the gradient information, an adversary cannot determine the source client of each fragment, nor can they reconstruct the original "gradient-like" vector from the fragmented components. This further prevents the adversary from identifying which items have zero gradients, thereby preserving user interaction privacy.

\section{Experiment}
\subsection{Experiment Settings}

\textbf{Datasets.} Four publicly accessible datasets obtained from various real-world online platforms are selected for experimentation: Amazon and Epinion. Their properties are summarized in Table \ref{tab:table1}. 

\begin{table}[h]
\caption{Information of all the datasets.}\label{tab:table1}
\centering
\begin{tabular*}{0.6\columnwidth}{@{\extracolsep{\fill}} ccccc @{}}
\toprule
No.  & Name      & |U|     & |I| & Density\\ 
\midrule
    D1      & Amazon    &	35,736  & 38,121	& 0.14\% \\  		      
    D2   & Epinion	  & 10,706	& 8,945	    & 0.31\% \\	
   
\bottomrule
\end{tabular*}
\end{table}

\textbf{Baselines.} In this study, we conduct a rigorous comparison between the proposed FHPLF model and eleven state-of-the-art baselines to highlight the advantages of binary representations in federated learning (FL). To evaluate the effectiveness of binary representations, we select DCF\cite{zhang2016discrete}, DPR\cite{zhang2017discrete}, CCCF\cite{liu2019compositional}, Neu-hash\cite{hansen2020content}, VHPHD\cite{hansen2021projected},  and HS-GCN\cite{liu2022hs} as centralized hash learning baselines. In addition,  PFedRec\cite{ijcai2023p507}, and RFRec\cite{liu2024efficient} are adopted as federated real-valued baselines, while LightFR\cite{zhang2023lightfr} is employed as a representative federated hash model. This comprehensive experimental design facilitates a systematic comparison between real-valued and binary representations across both centralized and federated learning settings, in terms of accuracy and efficiency.

\textbf{Evaluation Metrics.} Mean absolute error (MAE) and root mean squared error (RMSE) \cite{luo2021fast,wu2020data,wu2022double} are adopted to evaluate rating prediction accuracy. For ranking performance, we use hit ratio (HR), mean reciprocal rank (MRR), and normalized discounted cumulative gain (NDCG) \cite{wu2023hyperparameter}. Together, these metrics provide a comprehensive evaluation of both numerical prediction accuracy and ranking quality.

\subsection{Performance Comparison}
To comprehensively evaluate model performance, we conducted a systematic comparison with three categories of baselines: centralized hash-based models, federated real-valued models, and federated hash-based models.

\begin{table*}[ht]  
\centering
\caption{THE COMPARISON RESULTS ON PREDICTION OF RATING ACCURACY}\label{tab:table2}
\begin{adjustbox}{width=\textwidth}
\begin{tabular}{cc|cccccc|cc|cc}
\toprule
\multicolumn{2}{c}{Paradigms} &  \multicolumn{6}{c}{Centralized Hash} & \multicolumn{2}{c}{Federated Real value} & \multicolumn{2}{c}{Federated Hash} \\
\cmidrule(lr){1-2}\cmidrule(lr){3-8} \cmidrule(lr){9-10} \cmidrule(lr){11-12}
Dataset & Metrics 
& DCF & DPR & CCCF & Neu-hash & HS-GCN & VHPHD & PFedRec & RFRec & LightFR &  FHPLF \\
\midrule

\multirow{2}{*}{D1} 
& MAE  & 0.6717 & 0.9420 & 0.9397 & 0.6801 & 1.3361 & 0.9980 & 0.6661 & 0.6600 & 0.9424 & 0.6093 \\
& RMSE & 0.8538 & 1.1514 & 1.1902 & 0.8571 & 1.5195 & 1.1562 & 0.8684 & 0.8554  & 1.1011 & 0.8358 \\
\midrule

\multirow{2}{*}{D2} 
& MAE  & 0.9163 & 1.4890 & 1.4041 & 0.8934 & 1.5070 & 1.2384 & 0.8611 & 0.9146  & 1.0738  & 0.7835 \\
& RMSE & 1.1103 & 1.7079 & 1.6018 & 1.0975 & 1.6816 & 1.3959 & 1.1188 & 1.1450  & 1.2853  & 1.0973 \\

\bottomrule
\end{tabular}
\end{adjustbox}
\end{table*}

\begin{table*}[ht]  
\centering
\caption{THE COMPARISON RESULTS ON PREDICTION OF RANKING ACCURACY}\label{tab:table3}
\begin{adjustbox}{width=\textwidth}
\begin{threeparttable}

\begin{tabular}{cc|cccccc|cc|cc}
\toprule
\multicolumn{2}{c}{Paradigms} &  \multicolumn{6}{c}{Centralized Hash} & \multicolumn{2}{c}{Federated Real value} & \multicolumn{2}{c}{Federated Hash} \\
\cmidrule(lr){1-2}\cmidrule(lr){3-8} \cmidrule(lr){9-10} \cmidrule(lr){11-12}
Dataset & Metrics 
& DCF & DPR & CCCF & Neu-hash & HS-GCN & VHPHD & PFedRec & RFRec & LightFR &  FHPLF \\
\midrule

\multirow{3}{*}{D1} 
& HIT@10  & 0.9890 & 0.9137 & 0.9675 & 0.9896 & 0.9874 & 0.9817 & 0.9891 & 0.9893 & 0.9811 & 0.9903 \\
& MRR@10  & 0.6774 & 0.4706 & 0.5746 & 0.6842 & 0.6458 & 0.5877 & 0.6773 & 0.6840 & 0.5729 & 0.6931 \\
& NDCG@10 & 0.7551 & 0.5774 & 0.6710 & 0.7605 & 0.7311 & 0.6857 & 0.7552 & 0.7602 & 0.6743 & 0.7675 \\
\midrule

\multirow{3}{*}{D2} 
& HIT@10  & 0.9978 & 0.9841 & 0.9892 & 0.9980 & 0.9975 & 0.9930 & 0.9974 & 0.9974 & 0.9936 & 0.9981 \\
& MRR@10  & 0.8015 & 0.5564 & 0.6571 & 0.7956 & 0.7946 & 0.6172 & 0.7933 & 0.7691 & 0.6301 & 0.8135 \\
& NDCG@10 & 0.8515 & 0.6627 & 0.7401 & 0.8472 & 0.8463 & 0.7119 & 0.8453 & 0.8272 & 0.7218  & 0.8606 \\
\bottomrule
\end{tabular}

\end{threeparttable}
\end{adjustbox}

\end{table*}

\textbf{Compared with Centralized Hash-based Baselines:} We compare FHPLF with six representative centralized hash-based recommendation models, including DCF, DPR, CCCF, Neu-hash, HS-GCN, and VHPHD, using 64-bit hash codes for all methods. As shown in Tables \ref{tab:table2} and \ref{tab:table3}, FHPLF achieves the best overall performance on both datasets in terms of rating prediction and ranking recommendation. Specifically, FHPLF obtains the lowest MAE on D1 and D2, while achieving the best or highly competitive RMSE results. For ranking evaluation, FHPLF consistently achieves the highest HR@10, MRR@10, and NDCG@10 across both datasets, demonstrating the effectiveness of the proposed projected Hamming distance learning framework.

\textbf{Compared with Federated Learning Baselines:} Among federated learning-based models, we conduct a comprehensive comparison between the proposed FHPLF and three federated real-valued models (FedPerGNN, PFedRec, and RFRec), as well as a state-of-the-art federated hash-based model (LightFR). The experimental results demonstrate that FHPLF delivers consistently superior performance compared with federated learning models under prediction accuracy and communication efficiency.

\begin{table*}[h]
\caption{Properties of all Federated Learning models.}\label{tab:table4}
\centering
\begin{tabular*}{\textwidth}{@{\extracolsep{\fill}} c|ccccc|ccc @{}}
\toprule
Model  & Communication    & D1   & D2 & Privacy   & Inference & Accuracy  \\ 
\midrule
	
		PFedRec    & $O(|I|\times D \times 64)\uparrow \downarrow $ & 0.0364  & 0.0085   & Mid    &Slow	& Mid  \\	
        RFRec    & $O(|I|\times D \times 64)\uparrow \downarrow $    & 0.0364 & 0.0085    & Mid    &Slow	& Mid \\	
        LightFR    & $O(|I|\times D)\uparrow \downarrow $    & 0.0023  & 0.0005    & High	&Fast	&Mid  \\	
        FHPLF     & $O(|I|\times D)\uparrow \downarrow $     & 0.0023  & 0.0005   & High	&Fast	& High\\	
\bottomrule
\end{tabular*}
\end{table*}

Beyond prediction accuracy, we further compare the communication efficiency and practical properties of federated recommendation models. As shown in Table \ref{tab:table4}, hash-based methods (LightFR and FHPLF) require significantly lower communication costs than real-valued methods (PFedRec and RFRec). Specifically, the communication overhead is reduced from 0.0364 to 0.0023 on D1 and from 0.0085 to 0.0005 on D2, resulting in approximately 16--17$\times$ lower transmission cost. In addition, hash-based models provide faster inference and stronger privacy protection due to their compact binary representations. Among all federated methods, FHPLF achieves the same low communication cost and high privacy level as LightFR, while delivering higher recommendation accuracy. These results demonstrate that FHPLF effectively balances communication efficiency, privacy preservation, and recommendation performance in federated environments.

\subsection{Privacy Protection Evaluation}
To evaluate the privacy-preserving capability of FHPLF, we conduct gradient inversion attacks (GIA) and compare the results with the federated hashing model LightFR. Following the attack protocol, dummy user hash vectors and ratings are randomly initialized and iteratively optimized to reconstruct the original ratings from shared gradients.

\begin{table}[h]
\caption{Privacy evaluation of LightFR and FHPLF Under GIA.}\label{tab:table5}
\centering
\begin{tabular*}{0.7\linewidth}{@{\extracolsep{\fill}}cc|ccc}
\toprule
\multicolumn{1}{c}{Dataset} & \multicolumn{1}{c}{Metrics} & 
\multicolumn{1}{c}{LightFR} & \multicolumn{1}{c}{unSBG-PEU}  & \multicolumn{1}{c}{SBG-PEU} \\
\midrule

\multirow{2}{*}{D1}
& MAE     & 1.3150  &  1.4954  &  1.5020 \\
& RMSE    & 1.3454 &  1.8400  &  1.8449 \\

\multirow{2}{*}{D2}
& MAE     & 1.3088   & 1.5377    &  1.5403 \\
& RMSE    & 1.3402 &  1.8788  &  1.8826 \\

\bottomrule
\end{tabular*}
\end{table}

As shown in Table \ref{tab:table5}, LightFR achieves the lowest reconstruction errors on both datasets, indicating that the attacker can recover user ratings more accurately. In contrast, both FHPLF variants produce substantially larger MAE and RMSE values, making the recovered ratings less accurate and demonstrating stronger resistance to gradient inversion attacks. Specifically, FHPLF (SBG-PEU) achieves the highest reconstruction errors on D1 and D2, slightly outperforming FHPLF (unSBG-PEU) in privacy protection. These results indicate that the proposed SBG-PEU mechanism further enhances the privacy-preserving capability of FHPLF while maintaining effective recommendation performance.

\section{Conclusions}
In this paper, we proposed FHPLF, a federated hash-based latent factor framework that jointly addresses the challenges of communication efficiency, privacy preservation, and prediction accuracy in federated learning. By integrating projection Hamming distance learning and binary representation optimization, FHPLF reduces communication overhead while maintaining competitive recommendation performance. Furthermore, the proposed SBG-PEU mechanism enhances privacy protection against gradient inversion attacks. Experimental results on real-world datasets demonstrate that FHPLF consistently achieves superior performance in terms of accuracy, efficiency, and security compared with existing federated methods.

\bibliographystyle{unsrt}
\bibliography{references.bib} 

\end{document}